\documentclass[10pt,twocolumn,letterpaper]{article}

\usepackage[pagenumbers]{cvpr} 
\usepackage{color, colortbl}
\usepackage{multirow}
\usepackage{pifont}       
\usepackage{bbding}       
\usepackage{graphicx}
\usepackage{float}
\usepackage{subfloat}
\usepackage{xcolor}
\usepackage{color}
\definecolor{cvprblue}{rgb}{0.21,0.49,0.74}
\definecolor{skyblue}{RGB}{135,206,235}
\usepackage[pagebackref,breaklinks,colorlinks,allcolors=cvprblue]{hyperref}

\def\confYear{2026}

\title{Enhanced Partially Relevant Video Retrieval through \\Inter- and Intra-Sample Analysis with Coherence Prediction
}

\author{ Junlong Ren$^{\dag}$, Gangjian Zhang$^{\dag}$, 
        {Yu Hu$^{}$, Jian Shu$^{}$, Hui Xiong$^{}$, Hao Wang$^{}$\thanks{Corresponding Author. $^{\dag}$Equal Contribution.}} \\
        The Hong Kong University of Science and Technology (Guangzhou)\\
        \texttt{\{jren686,gzhang292,yhu847,jshu704\}@connect.hkust-gz.edu.cn},\\ 
        \texttt{\{xionghui,haowang\}@hkust-gz.edu.cn}}

\begin{document}
\maketitle
\begin{abstract}

Partially Relevant Video Retrieval (PRVR) aims to retrieve the target video that is partially relevant to the text query. The primary challenge in PRVR arises from the semantic asymmetry between textual and visual modalities, as videos often contain substantial content irrelevant to the query. Existing methods coarsely align paired videos and text queries to construct the semantic space, neglecting the critical cross-modal dual nature inherent in this task: inter-sample correlation and intra-sample redundancy. To this end, we propose a novel PRVR framework to systematically exploit these two characteristics. Our framework consists of three core modules. First, the Inter Correlation Enhancement (ICE) module captures inter-sample correlation by identifying semantically similar yet unpaired text queries and video moments, combining them to form pseudo-positive pairs for more robust semantic space construction. Second, the Intra Redundancy Mining (IRM) module mitigates intra-sample redundancy by mining redundant moment features and distinguishing them from query-relevant moments, encouraging the model to learn more discriminative representations. Finally, to reinforce these modules, we introduce the Temporal Coherence Prediction (TCP) module, enhancing discrimination of fine-grained moment-level semantics by training the model to predict the original temporal order of randomly shuffled video sequences. Extensive experiments demonstrate the superiority of our method, achieving state-of-the-art results.

\end{abstract}

\section{Introduction}

\begin{figure}[!t]
\centering
\includegraphics[width=1\linewidth]{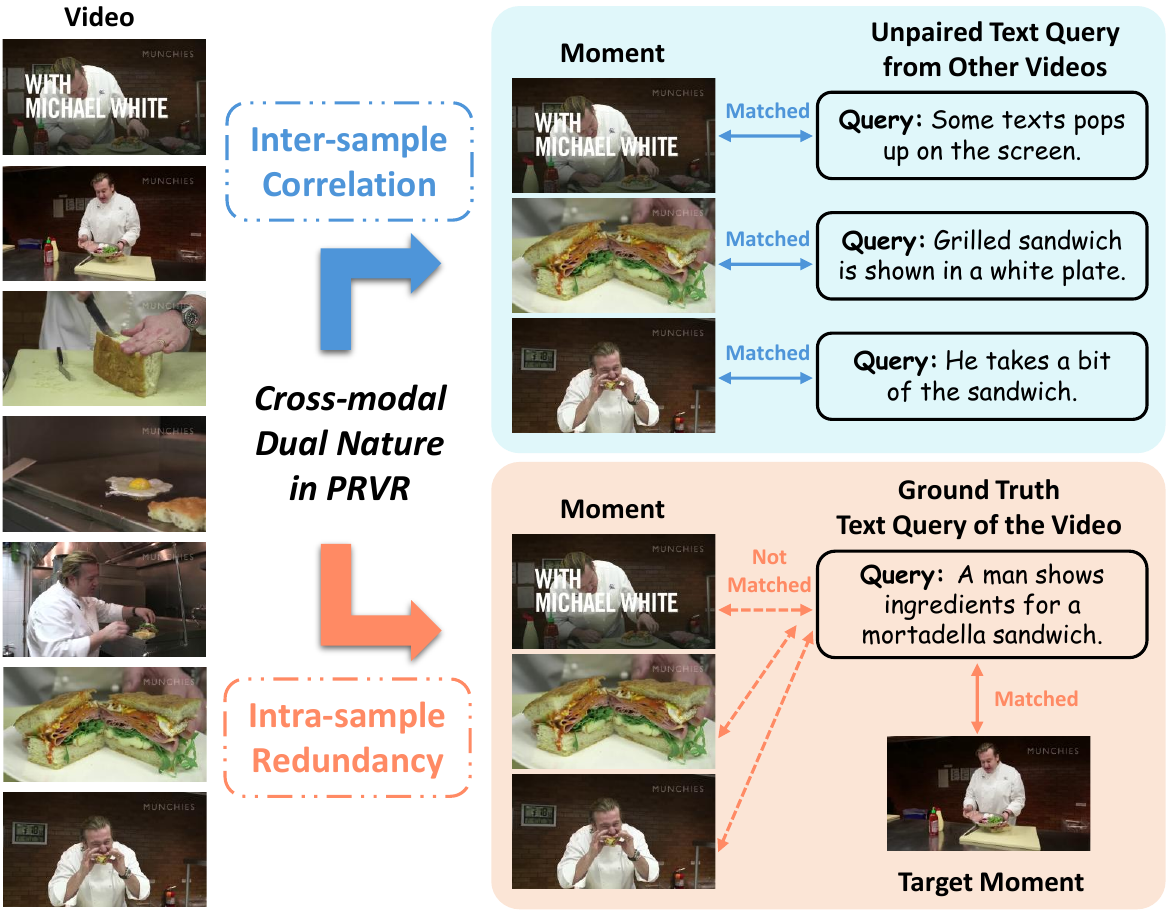}
\vspace{-8mm}
\caption{
In Partially Relevant Video Retrieval (PRVR), video-text samples exhibit the inherent cross-modal dual nature:
(a) \textbf{Inter-sample correlation}: The video contains certain moments that are semantically correlated to other unpaired text queries.
(b) \textbf{Intra-sample redundancy}: Apart from the target moment, other redundant moments in the video are irrelevant to the paired text query.
}
\label{fig::motivation}
\vspace{-4mm}
\end{figure}

With the explosion of online video content, users increasingly seek efficient ways to find specific videos within vast video collections. While conventional Text-to-Video Retrieval (T2VR) focuses on identifying pre-trimmed short clips that fully match a given text query, many real-world scenarios require identifying long videos that contain only partial moments relevant to a text query. This task is formally known as Partially Relevant Video Retrieval (PRVR).

The fundamental challenge in PRVR arises from the inherent semantics asymmetry between text and video modalities. While a video comprises multiple moments, only a subset may exhibit relevance to a given text query. Existing methods \cite{MSSL,PEAN,DKD,GMMFORMER,yin2024exploiting,MGAKD} approach this task from a limited perspective, primarily adopting conventional alignment constraints such as triplet ranking \cite{faghri2018vse++} and InfoNCE \cite{oord2018representation} to establish coarse associations between annotated paired text queries and target videos. These approaches may be effective when the semantic content of the two modalities is largely equivalent, as in the traditional T2VR. However, in PRVR, this design fails to distinguish the fine-grained moment events within the same video, leaving the rich semantics in videos unexplored.

Formally, these methods overlook the unique cross-modal dual nature in the PRVR task: namely inter-sample correlation and intra-sample redundancy. As illustrated in Figure~\ref{fig::motivation}, the video modality inherently shows richer semantics than the corresponding text query. Moments not explicitly related to the query may still contain valuable visual semantics that correlate with other unpaired queries in the dataset.
Conversely, the video modality also suffers from semantic redundancy, as only a limited subset of its content aligns with the given textual descriptions \cite{MSSL}. This results in a substantial portion of video content being irrelevant to the text query, which may not only fail to contribute to retrieval accuracy but could actively hinder the proper alignment between text queries and their corresponding video moments.
Moreover, videos inherently have rich temporal dynamics, where frame order has semantic significance. However, prior works lack explicit temporal coherence modeling, failing to discriminate fine-grained moment-level semantics.

In this paper, we introduce a novel framework for PRVR that systematically exploits the cross-modal dual nature of video-text relationships: inter-sample correlation and intra-sample redundancy, complemented by temporal coherence prediction to enhance temporal structure learning. The proposed framework comprises three key components: 
\textbf{(1)} The Inter Correlation Enhancement module is designed to fully exploit the inter-sample correlation of video-text modalities. It computes the cross-modal similarity between video moments and unpaired text queries. High-similarity pairs are identified and utilized as pseudo-positive samples, which are then incorporated into the alignment objective to construct a more comprehensive and discriminative cross-modal embedding space.
\textbf{(2)} To address intra-sample redundancy, the Intra Redundancy Mining module extracts semantically redundant video moments through joint analysis of text features, global video features, and local moment features. These redundant moments are distinguished from query-relevant moments during training, forcing the model to strengthen alignment between text and relevant video content and improve discrimination against irrelevant visual semantics within the same video.
\textbf{(3)} Complementing the above modules, the Temporal Coherence Prediction module enhances discrimination of fine-grained moment-level semantics through an auxiliary task. The model learns to predict the original temporal order of randomly shuffled video frames/moments, thereby developing a more robust understanding of video semantics and temporal structure.

Extensive experiments on three large-scale datasets, TVR \cite{tvr}, ActivityNet Captions \cite{activitynet}, and Charades-STA \cite{sta}, verify the superior performance and robustness of our method. Our contributions are summarized as follows:
\begin{itemize}
\item We introduce the Inter Correlation Enhancement module. It identifies high-similarity pairs between moments and unpaired text queries as pseudo-positive pairs, which are used to enrich cross-modal semantic space construction.
    
\item We propose the Intra Redundancy Mining module to extract redundant video moments and distinguish them from query-relevant moments. It forces the model to develop more precise and discriminative text-video alignment.

\item In the Temporal Coherence Prediction module, we design a self-supervised auxiliary task. It enhances discrimination of fine-grained moment-level semantics by predicting the original temporal order of shuffled video frames/moments. It results in more robust video features.
\end{itemize}

\section{Related Works}

\subsection{Partially Relevant Video Retrieval}
Partially Relevant Video Retrieval (PRVR) is to retrieve untrimmed videos that are only partially semantically related to the text query. Unlike previous video retrieval tasks, videos retrieved by PRVR do not necessarily exhibit complete semantic alignment with the textual query, often containing substantial redundant information. 
Specifically, MS-SL \cite{MSSL} is the first to propose the PRVR task. MS-SL learns video features at both the clip-scale and frame-scale.
GMMFormer \cite{GMMFORMER} increases information density in video encoding by adopting Gaussian windows. 
BGM-Net \cite{yin2024exploiting} proposes a new loss function to facilitate multimodal alignment based on unimodal similarities. 
However, these works overlook the unique cross-modal dual nature of PRVR, leading to limited semantic space construction. In contrast, we aim to exploit the inter-sample correlation and intra-sample redundancy, resulting in improved retrieval accuracy.

\begin{figure*}[!t]
\centering
\includegraphics[width=1\linewidth]{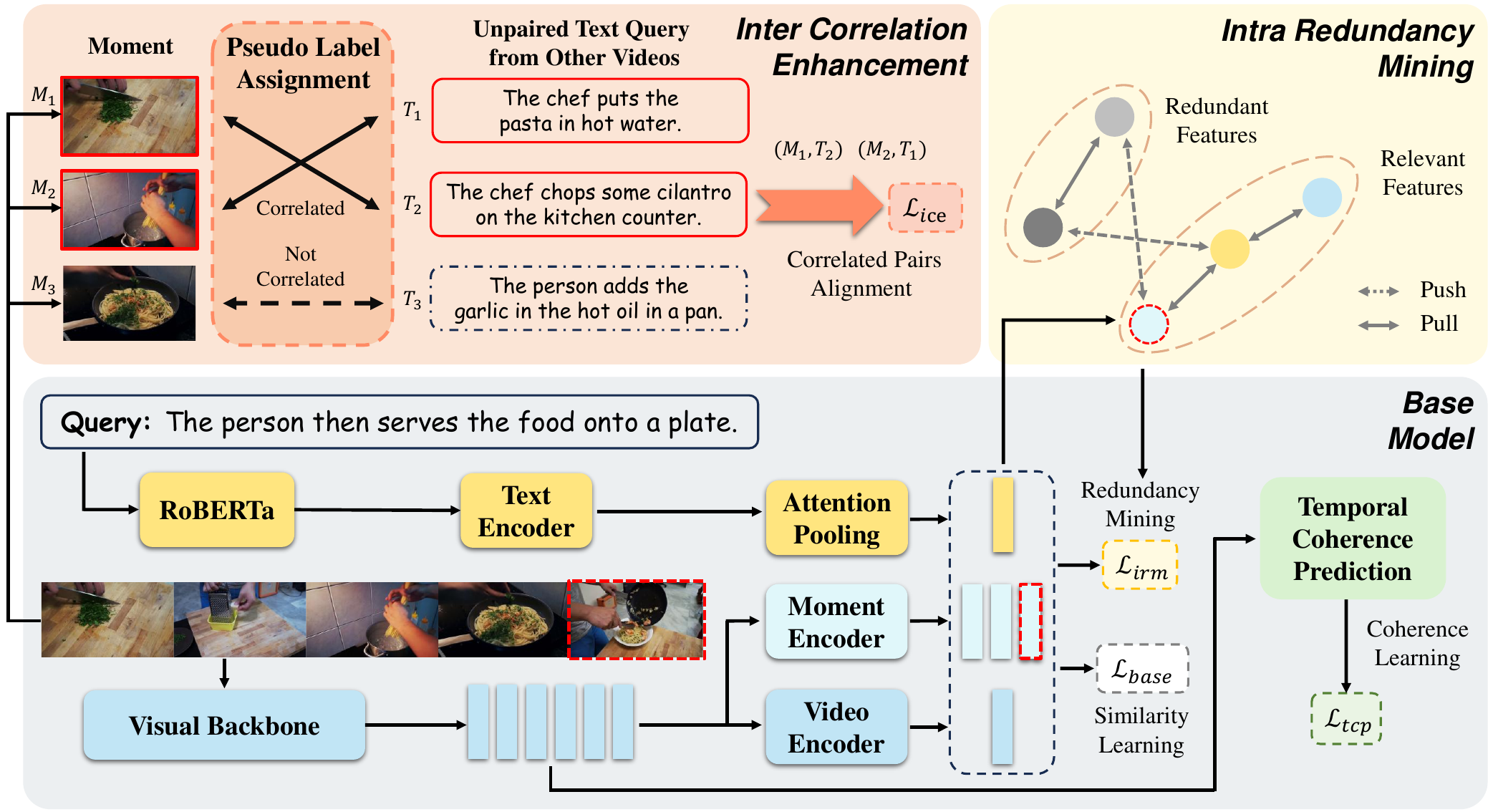}
\vspace{-6mm}
\caption{\textbf{Overview of the proposed framework.} We systematically leverage the cross-modal dual nature in PRVR, namely inter-sample correlation and intra-sample redundancy, complemented by temporal coherence prediction, to construct a more discriminative cross-modal semantic space. The framework comprises three key components: 
(a) \textbf{Inter Correlation Enhancement Module}: This component analyzes cross-modal correlations by identifying high-similarity pairs between video moments and unpaired text queries. These pseudo-positive pairs are incorporated during training to enrich the semantic space construction.
(b) \textbf{Intra Redundancy Mining Module}: The module extracts redundant video moments. By learning to distinguish these redundant moments and query-relevant moments, the model develops enhanced capability to focus on query-relevant visual semantics.
(c) \textbf{Temporal Coherence Prediction Module}: Designed to complement the other components, this module improves discrimination of fine-grained moment-level semantics through a self-supervised sequence prediction task, where the model predicts the original temporal order of shuffled video frames/moments.
}
\label{fig:overview}
\vspace{-4mm}
\end{figure*}

\subsection{Text-to-Video Retrieval}
In recent years, Text-to-Video Retrieval (T2VR) has attracted increasing attention. A common approach \cite{chen2020fine, li2019w2vv++, dong2019dual, dong2021dual,dong2022reading} involves retrieving pre-trimmed videos by comparing cross-modal similarities between textual queries and video content. Many studies \cite{wu2023cap4video,li2023unmasked,wang2024internvideo2} have proposed methods that project features from different modalities into a shared embedding space. However, these methods often neglect the fact that real-world videos may not consistently correspond to a single topic, and typically certain all moments within a video are relevant to the text query.

\section{Proposed Method}

\subsection{Problem Formulation}
\label{sec:Problem Formulation}
Partially relevant video retrieval (PRVR) aims to retrieve videos containing a moment semantically relevant to a text query, from a large corpus of untrimmed videos. 
Unlike conventional text-to-video retrieval, the videos are untrimmed and much longer, and text queries only correspond to a small portion of a video.
\textbf{Note that start/end timestamps of moments are unavailable in PRVR.}
The overview of our method is illustrated in Figure \ref{fig:overview}. 

\subsection{Base Model}
\label{sec:Base Model}

\subsubsection{Text Representation}
Given a sentence containing $N$ words, following previous works \cite{MSSL, GMMFORMER, yin2024exploiting}, we first employ the pre-trained RoBERTa \cite{roberta} to extract word embeddings. Then we utilize a fully-connected (FC) layer to project the word embeddings into a lower-dimensional space. 
We further adopt a standard Transformer layer \cite{TRANSFORMER} 
as the text encoder. The encoded word-level features are denoted as $Q = \left\{ q_{i} \right\}_{i = 1}^{N} \in \mathbb{R}^{N \times D}$, where $D$ is the feature dimension. Finally, we apply the additive attention mechanism \cite{bahdanau2014neural} on $Q$ to obtain the aggregated sentence-level feature $q \in \mathbb{R}^{D}$.

\subsubsection{Video Representation}
\label{sec:Video Representation}
Given an untrimmed video, we first adopt pre-trained CNN \cite{he2016deep, carreira2017quo} as the visual backbone to extract $T_f$ frame features. Following prior methods \cite{MSSL, GMMFORMER, yin2024exploiting,PEAN}, we employ a coarse-to-fine manner to capture the temporal information of videos by two distinct branches. The video-level branch models the global video-text similarity, while the moment-level branch measures the local moment-text similarity.

For the video-level branch, the frame features are fed into an FC layer, followed by the video encoder, which is a standard Transformer encoder. The encoded frame features are denoted as $V_f = \left\{ f_{i} \right\}_{i = 1}^{T_f} \in \mathbb{R}^{T_f \times D}$. Next, we employ the additive attention mechanism \cite{bahdanau2014neural} on $V_f$ to obtain the aggregated video-level feature $v \in \mathbb{R}^{D}$.

For the moment-level branch, we first condense the $T_f$ frame features into $T_m$ moment features by mean pooling over consecutive frame features. We then use an FC layer to reduce dimension. For the moment encoder, a standard Transformer encoder \cite{TRANSFORMER} is adopted to get contextual moment features $V_m = \left\{ m_{i} \right\}_{i = 1}^{T_m} \in \mathbb{R}^{T_m \times D}$.

\subsubsection{Similarity Learning}
\label{sec::Similarity Learning}
Given a video-text pair $\mathcal{V}$ and $\mathcal{T}$, we compute the similarity between the text feature and video features from the two aforementioned branches. Concretely, the video-level similarity $\mathcal{S}_v$ between the video feature $v$ and the sentence feature $q$ is measured by the cosine similarity:
\begin{equation}
    \mathcal{S}_v\left(\mathcal{V}, \mathcal{T}\right)=\operatorname{\textit{cos}}\left(v, q\right)=\frac{v^{\top} q}{\left\|v\right\| \cdot\left\|q\right\|}.
\end{equation}
Next, we compute the moment-level similarity $\mathcal{S}_m$ between moment features $V_m$ and sentence feature $q$, and identify the key moment feature $m_{k} \in \mathbb{R}^{D}$ with the highest similarity:
\begin{equation}
\label{key moment}
    k=\mathop{\arg\max}\limits_{i \in \{1, \cdots,T_m\}}(\operatorname{\textit{cos}}\left(m_{i}, q\right)),~\mathcal{S}_m\left(\mathcal{V}, \mathcal{T}\right)=\operatorname{\textit{cos}}\left(m_{k}, q\right).
\end{equation}

For cross-modal alignment, we adopt the combination of triplet ranking loss $\mathcal{L}^{trip}$ \cite{faghri2018vse++} and InfoNCE loss $\mathcal{L}^{nce}$ \cite{oord2018representation}, which is a common approach following prior works \cite{MSSL, PEAN, DKD, GMMFORMER, yin2024exploiting}.
The training loss of the base model is:
\begin{equation}
    \mathcal{L}_{base}=\mathcal{L}^{trip}_v+\mathcal{L}^{trip}_m+\lambda_1\mathcal{L}^{nce}_v+\lambda_2\mathcal{L}^{nce}_m,
\end{equation}
where $\mathcal{L}^{trip}_v$ and $\mathcal{L}^{trip}_m$ denote triplet losses using the video-level similarity $\mathcal{S}_v$ and moment-level similarity $\mathcal{S}_m$ as the cross-modal similarity, and accordingly for $\mathcal{L}^{nce}_v$ and $\mathcal{L}^{nce}_m$.
$\lambda_1$ and $\lambda_2$ are hyperparameters to balance losses.

\subsection{Inter Correlation Enhancement}
\label{sec:ICE}

\begin{figure}[!t]
\centering
\includegraphics[width=1\linewidth]{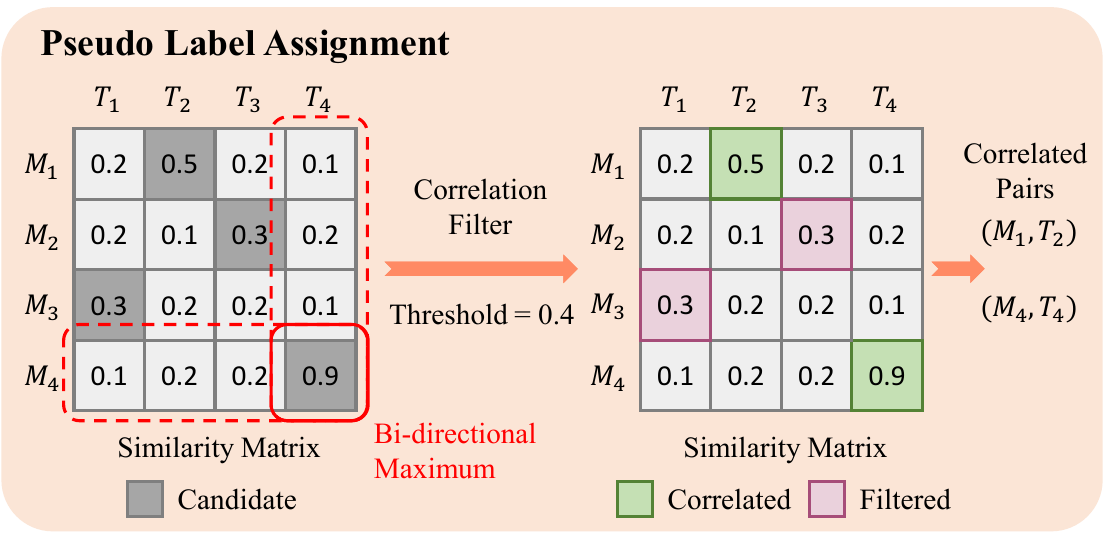}
\vspace{-6mm}
\caption{The ICE module employs a two-stage selection process for pseudo label assignment. 
It first computes the similarity between unpaired video moments and text features in a mini-batch. The pairs with mutual maximum similarity are selected as candidate pairs.
Then, only those pairs with a similarity higher than the threshold are retained, ensuring high-confidence pseudo labels.}
\label{fig::cae}
\vspace{-4mm}
\end{figure}

The PRVR task presents a fundamental semantic asymmetry, where textual queries only partially describe the content of target videos, leaving the visual modality inherently richer in semantic information. Our key observation reveals that while certain video moments may not align with their originally paired text annotations, they may exhibit strong semantic correspondence with other unpaired textual descriptions in the dataset, as demonstrated in Figure~\ref{fig::motivation}.
This inter-sample correlation represents a valuable but unexplored resource in current PRVR approaches. We propose to exploit these underlying cross-modal relationships.

To achieve this, we propose the Inter Correlation Enhancement (ICE) module.
Technically, we aim to learn these unlabeled but correlated visual-text relationships, thus constructing a richer semantic space. 
We exploit moment-text pairs that are most likely to be semantically corresponding while filtering out those semantically irrelevant ones. The pseudo label assignment process is shown in Figure~\ref{fig::cae}.

Given a mini-batch with $n$ video-text pairs, we first extract text and moment features as defined in Section \ref{sec:Base Model}. This results in $n_m=nT_m$ moment features and $n$ text features.
Then, we calculate a similarity matrix ${S} = \left\{ s_{ij} \right\} \in \mathbb{R}^{n_m \times n}$ for each moment feature and text feature.
Note that we manually set the similarities between each text and moments from its paired video to {-1} as we aim to mine unpaired correlated moment-text pairs.
For the $\bar{i}$-th moment and $\bar{j}$-th text, we find the most similar text and moment, respectively:
\begin{equation}
\begin{aligned}
    ~\hat{j}=\mathop{\arg\max}\limits_{j \in \{1, \cdots,n\}}(s_{\bar{i}j}),
    ~\hat{i}=\mathop{\arg\max}\limits_{i \in \{1, \cdots,n_m\}}(s_{i\bar{j}}),
\end{aligned}
\end{equation}
if $\hat{i}=\bar{i}$ and $\hat{j}=\bar{j}$, they are mutually the most similar to each other, and we deem them as a candidate pair.

Intuitively, a truly correlated pair should not only be the most similar to each other, but also exhibit high similarity.
To further filter out mismatched noisy pairs and enhance the accuracy of correlated pairs, only the pairs with a similarity higher than the correlation threshold of 0.4 are selected.

Finally, we obtain $n_c$ correlated pairs through the above two-stage selection.
These pairs then form a mini-batch of size $n_c$, and we align them using contrastive learning losses $\mathcal{L}^{trip}$ and $\mathcal{L}^{nce}$. The loss of ICE is denoted as $\mathcal{L}_{ice}$.

\subsection{Intra Redundancy Mining}
\label{sec:IRM}
The ICE module effectively harnesses inter-sample correlations to enhance the cross-modal semantic space for retrieval. However, while certain video semantics may align with unpaired text queries, they represent redundant information with respect to the originally paired text. This redundancy introduces noise during video-text matching, thereby limiting the retrieval accuracy.

To mitigate the intra-sample redundancy, we propose the Intra Redundancy Mining (IRM) module. It identifies redundant and query-irrelevant moment features by analyzing relations among text, video, and moment features. 
Unlike conventional negative samples (e.g., randomly sampled unpaired videos), these mined negatives exhibit higher feature similarity to the target moment, presenting a more challenging discrimination task. By forcing the model to distinguish redundant moments and query-relevant moments, we promote the learning of more discriminative fine-grained video representations, improving retrieval performance.

We construct redundant moment features from two perspectives. Technically, we model semantic variation in latent space between the key moment feature $m_{k}$ (Eq. \ref{key moment}), video feature $v$, and query feature $q$ to obtain video-view and query-view redundant features $r_v \in \mathbb{R}^{D}$ and $r_q \in \mathbb{R}^{D}$. Since semantic relationships are often captured as linear translations in latent space \cite{mikolov-etal-2013-linguistic, glove} and different modalities are mapped to the shared space with linear structure \cite{10418785}, we adopt subtraction to remove specific semantics \cite{TME}:
\begin{equation}
    r_v = FC(v-m_{k}),~r_q = FC(v-q),
\end{equation}
where $FC$ is a fully-connected layer to bridge the gap between video-moment and video-query differences, and $m_{k}$ is obtained in Eq.~\ref{key moment}. The intuition is that the video-level feature $v$ contains global semantics of all moments within the video. By subtracting the key moment feature $m_{k}$ from the video-level feature $v$, we obtain the video-view redundant feature $r_v$, which is deemed to be irrelevant to the text query. Similarly, we use a symmetrical operation to obtain the query-view redundant feature $r_q$, which is regarded as irrelevant to the target moment. 
Since both redundant features $r_v$ and $r_q$ should be semantically irrelevant to the key moment feature and query feature, we employ $r_v$ and $r_q$ as negative samples in the moment-level similarity learning (Section \ref{sec::Similarity Learning}), where the loss function is denoted as $\mathcal{L}_{neg}$.

Moreover, we propose to directly align $r_v$ and $r_q$. As the two redundant features are obtained at the video-view and query-view respectively, their alignment boosts the learning of underlying relationships between videos and text queries in the cross-modal semantic space. The corresponding loss $\mathcal{L}_{red}$ is also the jointly application of $\mathcal{L}^{trip}$ and $\mathcal{L}^{nce}$. The overall training objective of this module is the sum of the above two losses: $\mathcal{L}_{irm} = \mathcal{L}_{neg} + \mathcal{L}_{red}$.

\subsection{Temporal Coherence Prediction}
\label{sec:TCP}

\begin{figure}[!t]
\centering
\includegraphics[width=1\linewidth]{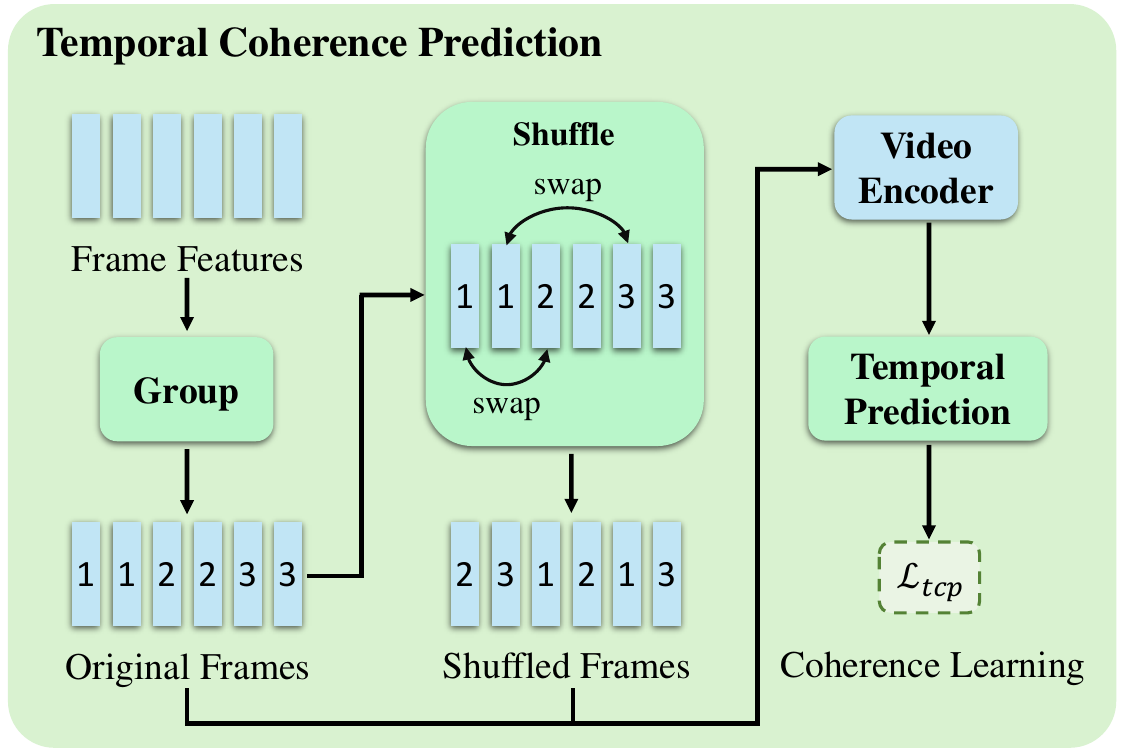}
\vspace{-6mm}
\caption{
The pipeline of the Temporal Coherence Prediction (TCP) module. The frame features are first divided into distinct groups. Then, a subset of frame features is randomly selected and shuffled, and the group labels of frames are predicted.
}
\label{fig::tcp}
\vspace{-4mm}
\end{figure}

While our ICE and IRM modules effectively leverage inter-sample correlations and mitigate intra-sample redundancy, their performance fundamentally depends on the model's temporal modeling ability. Clearer temporal feature distinctions enable more accurate cross-modal correlation measurement in ICE and more precise redundancy identification in IRM. To further enhance these capabilities, we introduce the Temporal Coherence Prediction (TCP) module, which strengthens discrimination of fine-grained moment-level semantics through self-supervised learning. 
Ablation studies in Sec. \ref{sec::Main Ablation Studies} indicate that TCP further enhances ICE and IRM.
The pipeline of TCP is shown in Figure \ref{fig::tcp}.

To capture the temporal structure of videos, we leverage the continuity between adjacent frames to divide the video sequence into coherent temporal segments. 
Inspired by Jigsaw Puzzle Solving in image representation learning \citep{wei2019iterative,jing2020self,zhuge2021kaleido}, we shuffle the video features to form an unordered sequence. The model is trained to predict the original temporal order. 
By explicitly modeling temporal coherence, this task helps the model to learn the coherence and causality between video events.
Instead of assigning extensive unique sequence labels, we adopt grouped labeling. Predicting unique labels in a long video sequence is overly hard and introduces noise. We address it by grouped labeling, reducing task complexity for robust temporal modeling. We also conduct ablation studies on group numbers in Section \ref{TCP}. 
Note that TCP is applied to both the video-level branch and moment-level branch in Section \ref{sec:Video Representation} to enable multi-grained temporal coherence learning. We mainly take the video-level branch for explanation. 

Concretely, we first divide frame features into $g$ groups, and each frame feature is given its group label ranging from $1$ to $g$. Every $T_f/g$ consecutive frame features are assigned the same group label.
Group labels of the original encoded frame features $V_f$ are denoted as $y_o \in \mathbb{R}^{T_f}$. 
We then randomly select and shuffle some frame features and encode them through the video encoder. 
Note that shuffling a small or large length of frames might lead to suboptimal performance. Too few shuffled frames provide insufficient learning signals, while excessive shuffling disrupts temporal coherence and makes the task overly challenging. 
Therefore, we only shuffle 25\% frames.
The shuffled frame features and their shuffled group labels are represented as $\hat{V_f} = \left\{ \hat{f_{i}} \right\}_{i = 1}^{T_f} \in \mathbb{R}^{T_f \times D}$ and $y_s \in \mathbb{R}^{T_f}$, respectively.
Then we adopt a classifier to predict the group labels of both the original and shuffled frame features $V_f$ and $\hat{V_f}$:
\begin{equation}
    p_o= Softmax(CLS(V_f)),~p_s= Softmax(CLS(\hat{V_f})),
\end{equation}
where $CLS$ is the classifier, $p_o \in \mathbb{R}^{T_f\times g}$ and $p_s \in \mathbb{R}^{T_f\times g}$ are the predicted probability distributions of group labels.
We adopt the cross-entropy loss to optimize this module:
\begin{equation}
    \mathcal{L}_{tcp} = f_{CE}(p_o, y_o) + f_{CE}(p_s, y_s),
\end{equation}
where $f_{CE}$ represents the cross-entropy loss function.

\subsection{Inference}
\label{sec:training objective}

During inference, the ICE, IRM, and TCP modules are deactivated.
The similarity of the video-text pair $\mathcal{V}$ and $\mathcal{T}$ is the weighted sum of moment-level and video-level similarities in Sec. \ref{sec::Similarity Learning}, following prior works \cite{MSSL,GMMFORMER, DKD, PEAN,yin2024exploiting}:
\begin{equation}
    \mathcal{S}\left(\mathcal{V}, \mathcal{T}\right)=\alpha~\mathcal{S}_m\left(\mathcal{V}, \mathcal{T}\right) + \left(1-\alpha\right) \mathcal{S}_v\left(\mathcal{V}, \mathcal{T}\right),
\end{equation}
where $\alpha \in \left[ 0, 1 \right]$ is a hyperparameter to balance similarities.

\begin{table*}[t]
\centering
\vspace{-2mm}
\caption{\textbf{Performance comparison results on TVR, ActivityNet Captions, and Charades-STA.} The best results are highlighted in \textbf{bold} and the second-best outcomes are \underline{underlined}. We achieve state-of-the-art results on all three datasets across all metrics.
* means built upon MS-SL \cite{MSSL}.
\textsuperscript{\dag} denotes using additional CLIP-B/32 model.
\textsuperscript{\ddag} indicates using Large Multimodal Models (LMMs).}
\label{table: comparison_with_sota}
\resizebox{1\linewidth}{!}{
\begin{tabular}{l|c|ccccc|ccccc|ccccc}
\toprule
\multirow{2.5}{*}{Method}&\multirow{2.5}{*}{Venue}&\multicolumn{5}{c|}{TVR} &\multicolumn{5}{c|}{ActivityNet Captions} &\multicolumn{5}{c}{Charades-STA} \\
\cmidrule(lr){3-7} \cmidrule(lr){8-12} \cmidrule(lr){13-17}
  & &R@1&R@5&R@10&R@100&SumR&R@1&R@5&R@10&R@100&SumR&R@1&R@5&R@10&R@100&SumR \\
 \midrule
 Cap4Video \cite{wu2023cap4video} & CVPR'2023 & 10.3 & 26.4 & 36.8 & 74.0 & 147.5 & 6.3 & 20.4 & 30.9 & 72.6 & 130.2 &  1.9 & 6.7 & 11.3 & 45.0 & 65.0 \\
 JSG~\cite{JSG} & MM'2023  & - & - & - & - & -   & 6.8 & 22.7 & 34.8 & 76.1 & 140.5   & 2.4 & 7.7 & 12.8 & 49.8 & 72.7 \\
 PEAN~\cite{PEAN} & ICME'2023 & 13.5 & 32.8 & 44.1 & 83.9 & 174.2  & 7.4 & 23.0 & 35.5 & 75.9 & 141.8    & {2.7} & {8.1} & {13.5} & 50.3 & {74.7} \\
  UMT-L~\cite{li2023unmasked} & ICCV'2023 & 13.7 & 32.3 & 43.7 & 83.7 & 173.4  & 6.9 & 22.6 & 35.1 & 76.2 & 140.8    & 1.9 & 7.4 & 12.1 & 48.2 & 69.6 \\
 InternVideo2\textsuperscript{\ddag}~\cite{wang2024internvideo2} & ECCV'2024 & 13.8 & 32.9 & 44.4 & 84.2 & 175.3  & 7.5 & 23.4 & 36.1 & 76.5 & 143.5    & 1.9 & 7.5 & 12.3 & 49.2 & 70.9 \\
 BGM-Net \cite{yin2024exploiting} & TOMM'2024 & 14.1 & 34.7 & {45.9} & {85.2} & {179.9} & 7.2 & 23.8 & 36.0 & 76.9 & 143.9 & 1.9 & 7.4 & 12.2 & 50.1 & 71.6 \\
 DL-DKD\textsuperscript{\dag}~\cite{DKD} & ICCV'2023 & {14.4} & {34.9} & 45.8 & 84.9 & {179.9}  & 8.0 & {25.0} & {37.5} & {77.1} & {147.6}    & - & - & - & - & - \\
 RAL*~\cite{RAL} & EMNLP'2025 & 14.5 & 34.3 & 45.8 & 84.5 & 179.1 & 7.4 & 23.4 & 35.4 & 76.7 & 143.0 & - & - & - & - & - \\
 PBU~\cite{PBU} & ICCV'2025 & 15.4 & 35.9 & 47.5 & 86.4 & 185.1 & 7.9 & 24.9 & 37.2 & 77.3 & 147.4 & - & - & - & - & - \\
 SDM*~\cite{SDM} & AAAI'2025 & 15.7 & 36.7 & 47.9 & 85.8 & 186.2 & 7.2 & 23.5 & 35.8 & 76.9 & 143.4 & - & - & - & - & - \\
 HLFormer~\cite{HLFormer} & ICCV'2025 & 15.7 & 37.1 & 48.5 & 86.4 & 187.7 & {8.7} & {27.1} & {40.1} & {79.0} & {154.9} & 2.6 & {8.5} & {13.7} & {54.0} & {78.7}\\
 MGAKD\textsuperscript{\dag}~\cite{MGAKD} & TOMM'2025 & {16.0} & {37.8} & {49.2} & {87.5} & {190.5} & 7.9 & {25.7} & {38.3} & {77.8} & {149.6} & - & - & - & - & - \\
 MS-SL++~\cite{MS-SL++} & TPAMI'2025 & 13.6 & 33.1 & 44.2 & 83.5 & 174.5 & 7.0 & 23.1 & 35.2 & 75.8 & 141.1 & 1.8 & 7.6 & 12.0 & 48.4 & 69.7\\
 \midrule
 MS-SL~\cite{MSSL} & MM'2022 & 13.5 & 32.1 & 43.4 & 83.4 & 172.4  & 7.1 & 22.5 & 34.7 & 75.8 & 140.1    & 1.8 & 7.1 & 11.8 & 47.7 & 68.4 \\
 \rowcolor{skyblue!40}\textbf{+ Ours} & - & 17.1	&37.9	&48.2	&86.2&	189.4&	9.8	&28.1	&41.1&	78.6	&157.6&
{2.4} & {9.1} & {14.6} & {53.9} & {80.0}  \\
 \midrule
 GMMFormer~\cite{GMMFORMER} & AAAI'2024 & 13.9 & 33.3 & 44.5 & 84.9 & 176.6  & {8.3} & 24.9 & 36.7 & 76.1 & 146.0    & 2.1 & 7.8 & 12.5 & {50.6} & 72.9 \\
 \rowcolor{skyblue!40}\textbf{+ Ours} & - & \textbf{18.6}&	\textbf{40.3}	&\textbf{51.1}&	\textbf{89.3}&	\textbf{199.3}&	\textbf{11.5}&	\textbf{30.2}&	\textbf{43.8}	&\textbf{81.6}&	\textbf{167.1}& \textbf{3.0} & \textbf{9.5} & \textbf{15.3} & \textbf{55.1} & \textbf{82.9}  \\
 \midrule
 Base Model & - &13.8	&32.2&	44.1	&83.7	&173.8&	7.0	&22.9	&35.1	&76.3&	141.3&	1.5&	6.8&	11.3&	49.5	&69.1\\
\rowcolor{skyblue!40}
 \textbf{+ Ours} & - & \underline{17.5} & \underline{39.0} & \underline{49.9} & \underline{87.6} & \underline{194.0} & \underline{10.1} & \underline{28.6} & \underline{41.9} & \underline{79.8} & \underline{160.4}     & \underline{2.9} & \underline{9.2} & \underline{14.9} & \underline{54.3} & \underline{81.3} \\
\bottomrule
\end{tabular}
}
\vspace{-3mm}
\end{table*}	

\section{Experiments}

\subsection{Experimental Setup}

\subsubsection{Datasets}
To evaluate the effectiveness of our method, we utilize three widely-used datasets with untrimmed long videos: TVR \cite{tvr}, ActivityNet Captions \cite{activitynet}, and Charades-STA \cite{sta}. 
\textbf{Note that we do not use moment-level annotations provided by these datasets in the setting of PRVR.}
Unlike conventional T2VR datasets \cite{xu2016msr, chen2011collecting, wang2019vatex} using pre-trimmed short clips, PRVR employs untrimmed long videos where the content is only partially relevant to text queries.
\textbf{TVR} is collected from six TV shows, containing 21.8K videos and 109K text queries. Each video has five textual descriptions on average, depicting different moments in the video. Following \cite{MSSL}, the training set consists of 17,435 videos with 87,175 moments, while the test set contains 2,179 videos with 10,895 moments.
\textbf{ActivityNet Captions} consists of a wide range of human activities, sourced from 20K YouTube videos. On average, each video has 3.7 moments with sentence descriptions. We adopt the same data partition as in \cite{MSSL, DKD,GMMFORMER}, where 10,009 and 4,917 videos (37,421 and 17,505 moments) are utilized as the training and test sets.
\textbf{Charades-STA} primarily comprises videos that depict indoor activities, including 6,670 videos and 16,128 moment annotations. We adopt the official data partition, where 12,408 and 3,720 moment-sentence pairs are used for training and evaluation, respectively.

\subsubsection{Evaluation Metrics}
Following previous works \cite{MSSL, DKD, GMMFORMER}, we adopt the rank-based metrics Recall Rate at $k$ (R@$k$) and the sum of all recall rates (SumR). R@$k$ measures the proportion of relevant items successfully retrieved within the top-$k$ results, with $k$ set to \{1, 5, 10, 100\}. Higher recall rates indicate better retrieval accuracy.

\subsubsection{Implementation Details}
For fair comparison with prior works on PRVR \cite{MSSL, PEAN, GMMFORMER, yin2024exploiting, DKD, MGAKD, PBU,HLFormer}, we utilize the same visual and textual features provided by \cite{MSSL}. The visual and textual features are extracted by pre-trained I3D \cite{carreira2017quo} with ResNet152 \cite{he2016deep} and RoBERTa \cite{roberta}.
The feature dimension $D$ is set to 384. $\lambda_1$, $\lambda_2$, $T_m$, and $\alpha$ are set to 0.02, 0.04, 32, and 0.7 following \cite{MSSL,GMMFORMER}.
$g$ is set to 8.
For training, we utilize the Adam optimizer \cite{KingmaB14} with an initial learning rate of 2.5e-4 and a mini-batch size of 128. The model is trained for 100 epochs with an early stopping strategy.
All experiments are conducted on a single NVIDIA RTX 3090 GPU.

\subsection{Comparison with State-of-the-Arts}

\subsubsection{Compared Methods}
We compare with other methods on TVR, ActivityNet Captions, and Charades-STA. These methods include PRVR \cite{MSSL, PEAN, GMMFORMER, yin2024exploiting, DKD, MGAKD, PBU,HLFormer,SDM,RAL}, T2VR \cite{wu2023cap4video,li2023unmasked}, and large multimodal models \cite{wang2024internvideo2}. 
For fair comparison, all compared PRVR methods adopt the same I3D/ResNet/RoBERTa features provided by \cite{MSSL}.

\subsubsection{Performance}
As summarized in Table \ref{table: comparison_with_sota}, our method achieves state-of-the-art (SOTA) results on all three datasets across all metrics. 
Notably, we outperform the previous SOTA method MGAKD \cite{MGAKD} without using additional CLIP models. 
Besides, we also surpass the Large Multimodal Model (LMM) InternVideo2 \cite{wang2024internvideo2} by a significant margin. InternVideo2 merely considers the similarity between text queries and entire videos, neglecting the essential fine-grained understanding of moment events. In contrast, our method fully leverages the rich semantics of distinct moment events and learns to distinguish them.
Comparison highlights the advantages and expertise of our method in the PRVR task.

To demonstrate the generalization of our method, we also integrate it into two existing PRVR models: MS-SL \cite{MSSL} and GMMFormer \cite{GMMFORMER}. Concretely, we directly replace the base model with these models.
As shown in Table \ref{table: comparison_with_sota}, our approach acts as a plug-and-play enhancement to these models on all datasets. 
The universal gains also highlight that our method is model-agnostic and effectively constructs a more comprehensive cross-modal semantic space.

\subsubsection{Efficiency}
We evaluate the efficiency of our proposed model and existing models using official codes. For fair comparison, all experiments are conducted using the same batch size and a single RTX 3090 GPU on ActivityNet Captions. All models are equipped with early stopping.
As shown in Table \ref{table::Efficiency}, with the same  I3D/RoBERTa features and fewer learnable parameters, we achieve faster training/inference speed and better retrieval accuracy. These results demonstrate the effectiveness and efficiency of our proposed model.

\begin{table}[t]     
\scriptsize
\centering
\caption{Efficiency comparison on ActivityNet Captions using the same I3D and RoBERTa features under identical conditions. 
}
\label{table::Efficiency}
\setlength{\tabcolsep}{2.5pt}
\resizebox{1\linewidth}{!}{
\begin{tabular}{l|c|c|c|cc}
\toprule
{Method} & {Model Size} & {Training Time} & {Inference Time} & {SumR} \\ \midrule
MS-SL \cite{MSSL} & \textbf{3.37M} & 166.05 min & 146.63s & 140.1 \\
GMMFormer \cite{GMMFORMER} & 11.37M & 89.05 min & 87.69s & 146.0 \\
SDM \cite{SDM} & 25.73M & 274.38 min & 91.85s & 156.6 \\
HLFormer \cite{HLFormer}  & 28.43M & 189.12 min & 85.39s & 154.9 \\
\midrule
\rowcolor{skyblue!40} \textbf{Ours} & 4.65M & \textbf{80.56 min} & \textbf{83.01s} & \textbf{160.4} \\
\bottomrule
\end{tabular}
}
\end{table}

\subsection{Ablation Studies}

\subsubsection{Main Ablation Studies}
\label{sec::Main Ablation Studies}

\begin{table}[t]     
\scriptsize
\centering
\caption{{Main ablation studies of the proposed key modules.}}
\label{table::ablation study all}
\setlength{\tabcolsep}{2.5pt}
\newcolumntype{C}[1]{>{\centering\arraybackslash}p{#1}}
\resizebox{1\linewidth}{!}{
\begin{tabular}{c|ccc|cc|C{0.95cm}C{0.95cm}|cc}
\toprule
\multirow{2.5}{*}{Row} & \multicolumn{3}{c|}{Setting} & \multicolumn{2}{c|}{TVR} & \multicolumn{2}{c|}{ActivityNet Captions} & \multicolumn{2}{c}{Charades-STA} \\
\cmidrule(lr){2-4} \cmidrule(lr){5-6} \cmidrule(lr){7-8} \cmidrule(lr){9-10}
&\textit{ICE}   & \textit{IRM} & \textit{TCP} &R@1&R@5&R@1&R@5&R@1&R@5\\ 
 \midrule
1 & \ding{56} & \ding{56} & \ding{56} & 13.8 & 32.2 & 7.0 & 22.9 & 1.5 & 6.8\\
\midrule
2 & \ding{51} & \ding{56} & \ding{56} & 14.6 & 33.9 & 7.8 & 24.9 & 2.2 & 7.5\\
3 & \ding{56} & \ding{51} & \ding{56} & 14.4 & 33.7 & 7.7 & 24.5 & 2.1 & 7.8\\
4 & \ding{56} & \ding{56} & \ding{51} & 14.1 & 33.5 & 7.4 & 24.1 & 1.7 & 7.0\\
\midrule
5 & \ding{56} & \ding{51} & \ding{51} & 15.7 & 36.5 & 8.1 & 26.8 & 2.4 & 8.3\\
6 & \ding{51} & \ding{56} & \ding{51} & 15.4 & 36.2 & 8.3 & 26.4 & 2.5 & 8.5\\
7 & \ding{51} & \ding{51} & \ding{56} & 16.0 & 36.9 & 9.1 & 27.0 & 2.6 & 8.8\\
\midrule
\rowcolor{skyblue!40} 8 & \ding{51} & \ding{51} & \ding{51} & \textbf{17.5} & \textbf{39.0} & \textbf{10.1} & \textbf{28.6} & \textbf{2.9} & \textbf{9.2}\\
\bottomrule
\end{tabular}
}
\vspace{-2mm}
\end{table}

\begin{table}[t]     
\scriptsize
\centering
\caption{Ablation study of loss terms in the IRM module.}
\label{table::ablation study rp}
\setlength{\tabcolsep}{2.5pt}
\newcolumntype{C}[1]{>{\centering\arraybackslash}p{#1}}
\resizebox{1\linewidth}{!}{
\begin{tabular}{c|cc|cc|C{0.95cm}C{0.95cm}|cc}
\toprule
\multirow{2.5}{*}{Row} & \multicolumn{2}{c|}{Loss Terms} & \multicolumn{2}{c|}{TVR} & \multicolumn{2}{c|}{ActivityNet Captions} & \multicolumn{2}{c}{Charades-STA} \\
\cmidrule(lr){2-3} \cmidrule(lr){4-5} \cmidrule(lr){6-7} \cmidrule(lr){8-9}
& $\mathcal{L}_{neg}$   & $\mathcal{L}_{red}$ &R@1&R@5&R@1&R@5&R@1&R@5\\ 
 \midrule
1 & \ding{56} & \ding{56}  & 15.4 & 36.2 & 8.3 & 26.4 & 2.5 & 8.5 \\
\midrule
2 & \ding{51} & \ding{56}  & 16.4 & 38.1 & 9.2 & 27.5 & 2.7 & 8.8  \\
3 & \ding{56} & \ding{51}  & 15.9 & 37.1 & 9.1 & 27.8 & 2.6 & 8.8  \\
\midrule
\rowcolor{skyblue!40} 4 & \ding{51} & \ding{51}  &\textbf{17.5} & \textbf{39.0} & \textbf{10.1} & \textbf{28.6} & \textbf{2.9} & \textbf{9.2}  \\
\bottomrule
\end{tabular}
}
\end{table}

To comprehensively validate the effectiveness of each module in our method, we conduct extensive ablation studies in Table \ref{table::ablation study all}. 
Comparing the results in each row, we have the following observations: 

\textbf{(1)} Each module independently improves performance over the base model (Row 1-4), suggesting their effectiveness in refining the cross-modal semantic space.

\textbf{(2)} Combining any two modules yields further improvements (Row 5-7), indicating their complementary roles in aligning semantics. Specifically, the joint application of ICE and IRM achieves the largest gain (Row 7), highlighting the necessity of exploiting the cross-modal dual nature of video-text modalities in PRVR.

\textbf{(3)} Integrating all three modules leads to the best performance (Row 8), with significant gains across all metrics. 
In particular, TCP further improves ICE and IRM. ICE and IRM improve R@1 by 2.2 on TVR (Row 1 and 7, 13.8 → 16.0). When combined with TCP, the performance gain increases to 3.4 (Row 4 and 8, 14.1 → 17.5).

\subsubsection{Intra Redundancy Mining}
We verify the effectiveness of each loss function in the Intra Redundancy Mining (IRM) module. As shown in Table \ref{table::ablation study rp}, 
both $\mathcal{L}_{neg}$ (Row 2) and $\mathcal{L}_{red}$ (Row 3) lead to boosts on all metrics.
Moreover, combining $\mathcal{L}_{neg}$ and $\mathcal{L}_{red}$ (Row 4) yields the highest results. The effectiveness arises because $\mathcal{L}_{neg}$ directly suppresses redundancy by pushing away redundant and query-relevant moments, while $\mathcal{L}_{red}$ implicitly strengthens cross-modal alignment by correlating video-view and query-view redundancies.

\begin{table}[t]     
\scriptsize
\centering
\caption{Ablations of the group number $g$ in the TCP module.}
\label{table::ablation group number}
\setlength{\tabcolsep}{3.5pt}
\newcolumntype{C}[1]{>{\centering\arraybackslash}p{#1}}
\resizebox{0.95\linewidth}{!}{
\begin{tabular}{c|c|cc|C{0.95cm}C{0.95cm}|cc}
\toprule
\multirow{2.5}{*}{Row} & \multirow{2.5}{*}{$g$} & \multicolumn{2}{c|}{TVR} & \multicolumn{2}{c|}{ActivityNet Captions} & \multicolumn{2}{c}{Charades-STA} \\
\cmidrule(lr){3-4} \cmidrule(lr){5-6} \cmidrule(lr){7-8} 
& &R@1&R@5&R@1&R@5&R@1&R@5\\
\midrule
1 & 4 & 17.1 & 38.4 & 9.7 & 28.1 & 2.7 & 9.0 \\\midrule
\rowcolor{skyblue!40} 2 & \textbf{8} & \textbf{17.5} & \textbf{39.0} & \textbf{10.1} & \textbf{28.6} & \textbf{2.9} & \textbf{9.2} \\ \midrule
3 & 16 & 16.4 & 37.2 & 9.2 & 27.4 & 2.4 & 8.6 \\
4 & 32 & 15.7 & 36.8 & 8.4 & 26.7 & 2.2 & 7.9 \\ \bottomrule
\end{tabular}
}
\vspace{-2mm}
\end{table}

\begin{figure*}[t]
\centering
\includegraphics[width=1\linewidth]{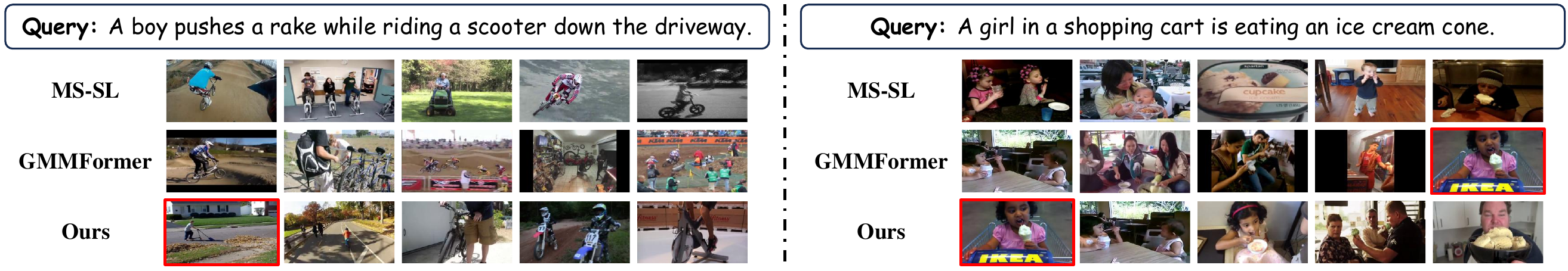}
\vspace{-6mm}
\caption{Visualization comparisons of retrieval results between our method, MS-SL \cite{MSSL}, and GMMFormer \cite{GMMFORMER} on ActivityNet Captions. 
The top-5 retrieval results are shown from left to right.
Ground-truth videos are marked with \textcolor[rgb]{1,0,0}{red boxes}.
Zoom in for better visibility.
}
\vspace{-1mm}
\label{fig::retrieval_result}
\end{figure*}

\begin{figure}[t]
\centering
\subfloat[Base Model]{
\includegraphics[width=0.48\columnwidth]{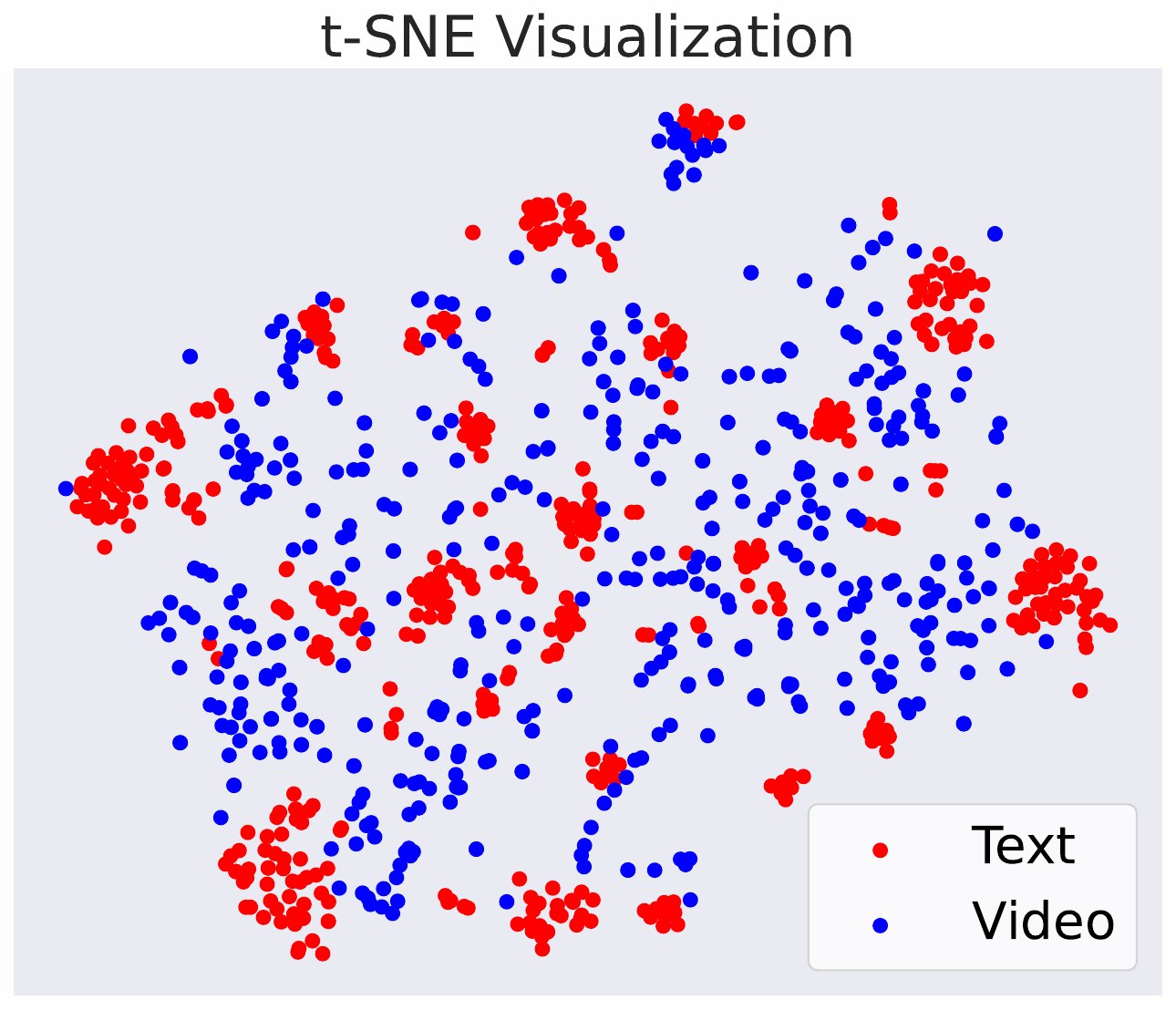}
\hspace{-0.em}}
\subfloat[Full Model (Ours)]{
\includegraphics[width=0.48\columnwidth]{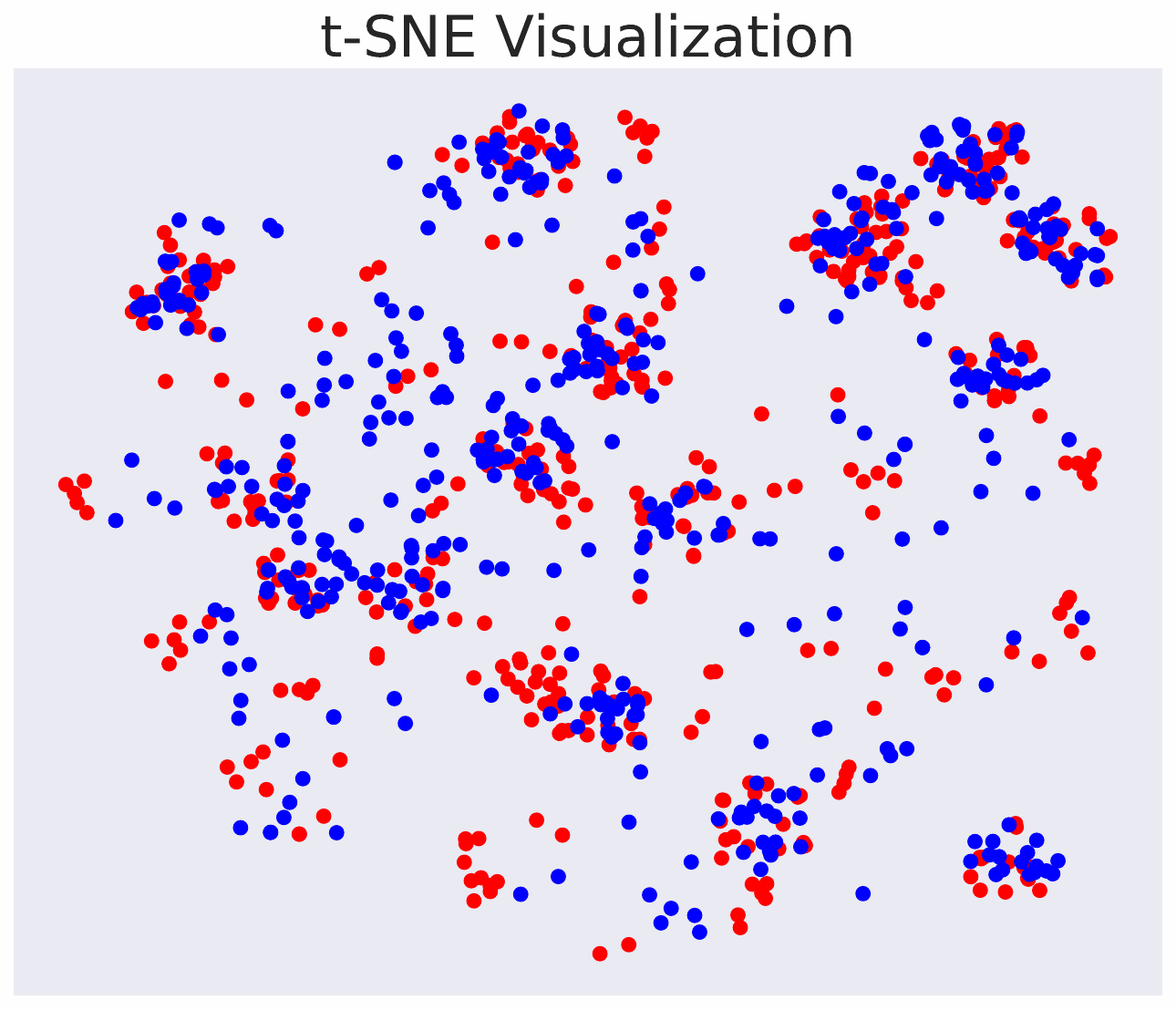}
}
\vspace{-3mm}
\caption{t-SNE visualization \cite{tsne} of text and video features on Charades-STA.
(a) is the base model trained with only $\mathcal{L}_{base}$.
(b) shows the full model trained with the full setup.
}
\label{fig:tsne}
\end{figure}

\begin{figure}[t]
\centering
\subfloat{\includegraphics[width=0.33\columnwidth]{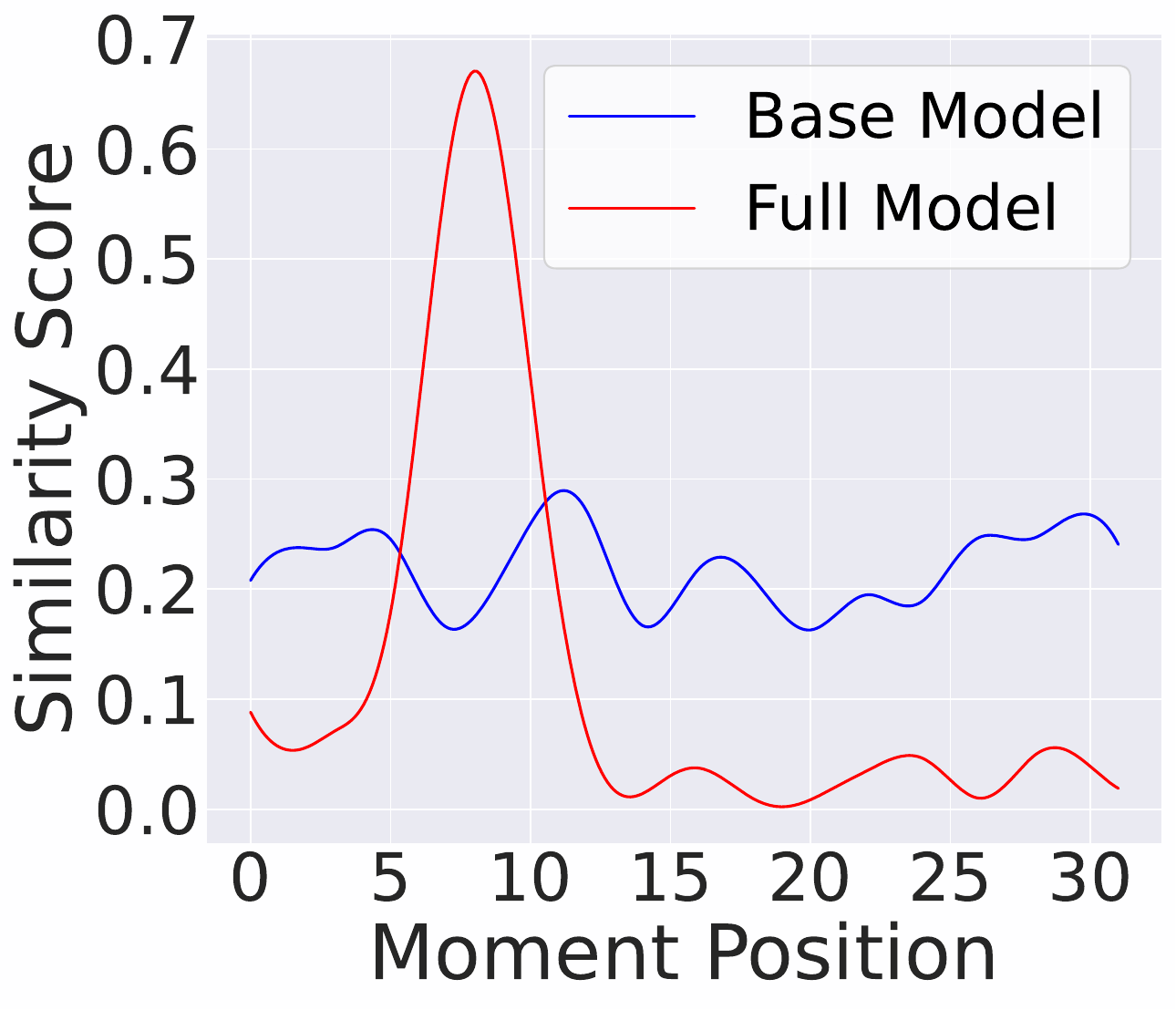}\hspace{-0.em}}
\subfloat{\includegraphics[width=0.33\columnwidth]{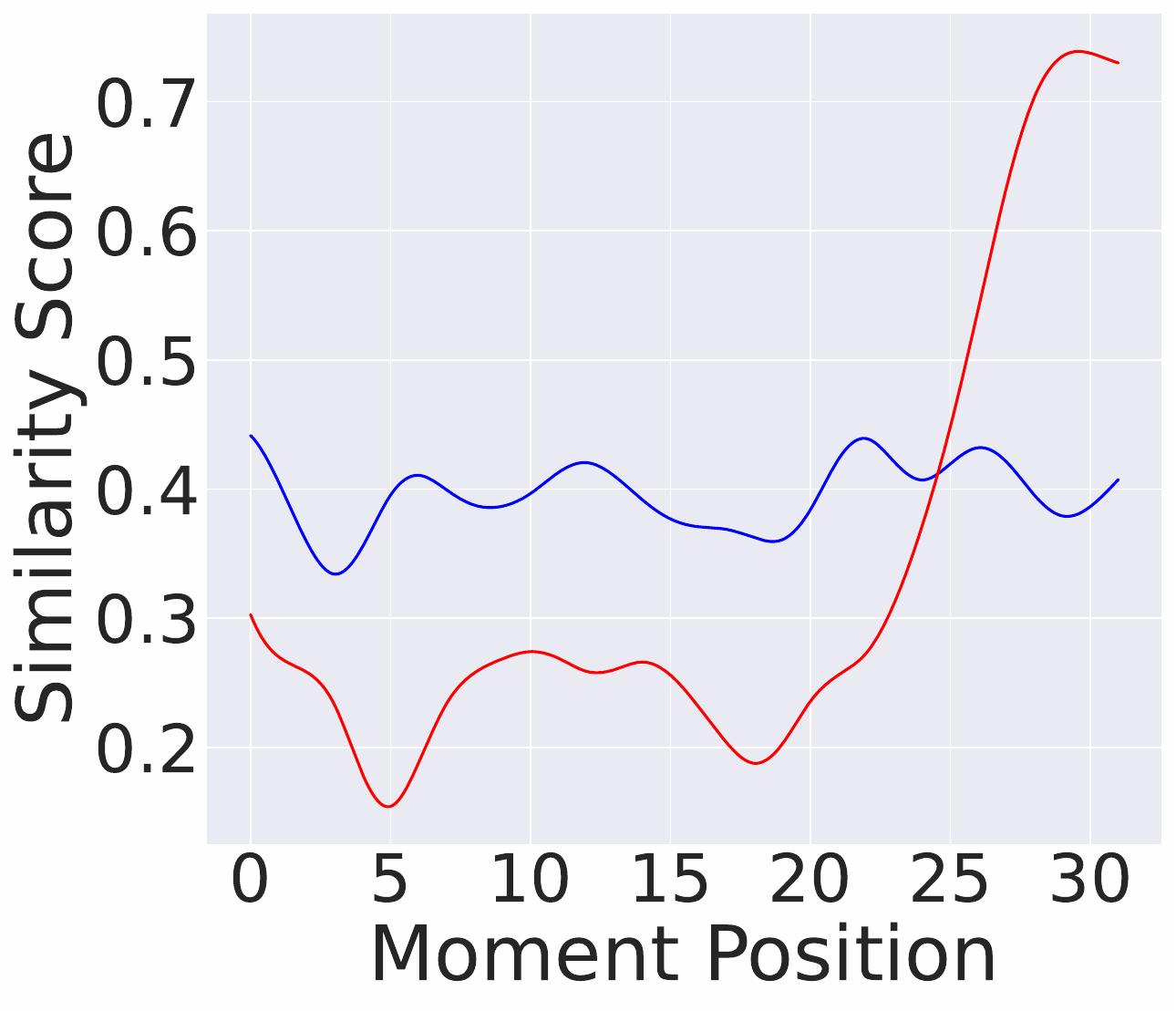}\hspace{-0.em}}
\subfloat{\includegraphics[width=0.33\columnwidth]{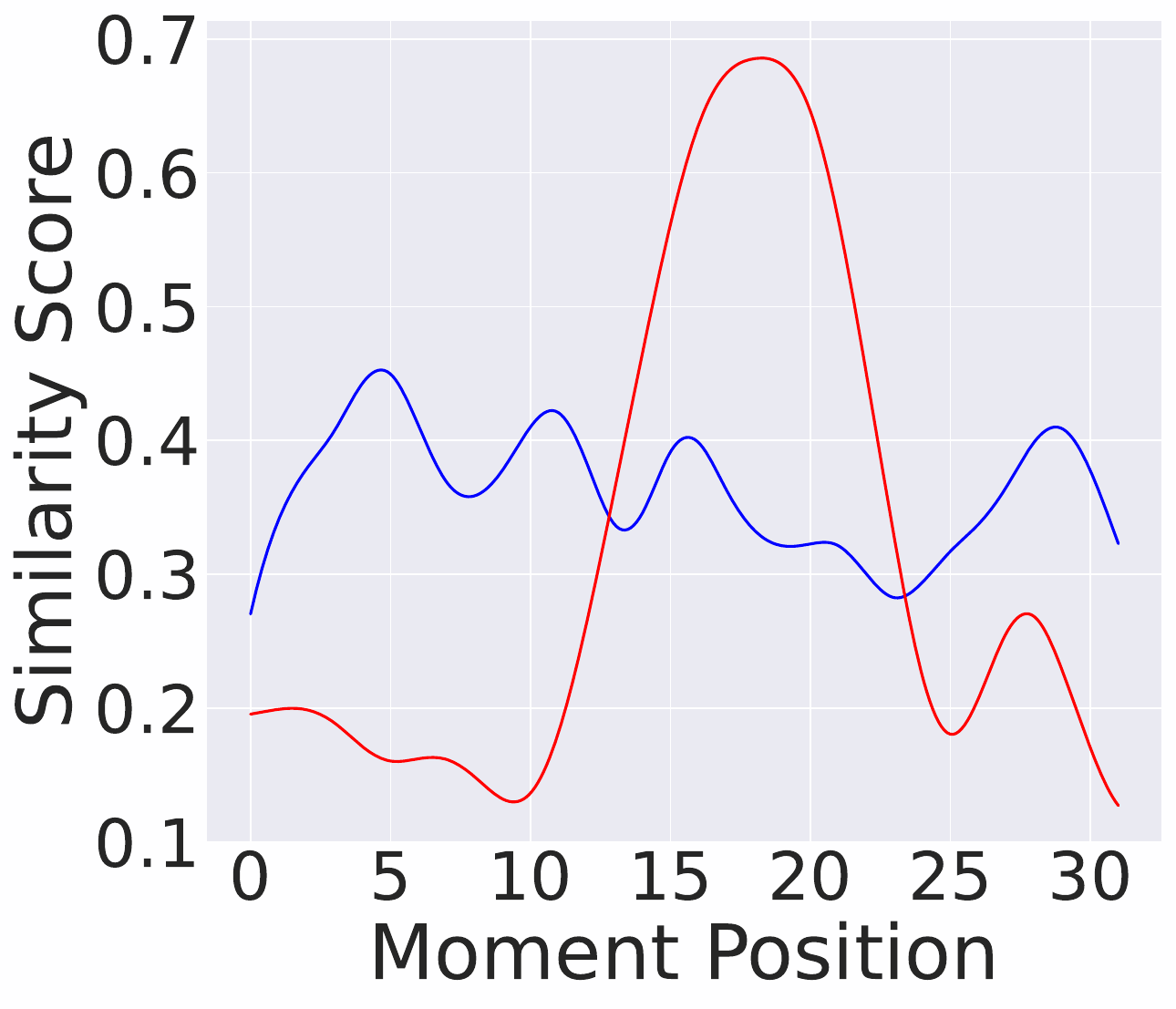}}
\vspace{-2mm}
\caption{Text-moment similarity on TVR.
We smooth out similarity intervals for better readability.
}
\label{fig:Text-Moment Similarity}
\vspace{-2mm}
\end{figure}

\subsubsection{Temporal Coherence Prediction}  
\label{TCP}

We assess the group number $g$ in the Temporal Coherence Prediction (TCP) module, which controls the granularity of temporal coherence learning. As shown in Table \ref{table::ablation group number}, performances decline as $g$ $>$ 8 in all datasets due to increased task difficulty. 
The optimal results are achieved at $g$ = 8, suggesting that moderate grouping balances task difficulty and learning efficacy.

\subsection{Qualitative Results}

\subsubsection{Retrieval Results}
To qualitatively validate the effectiveness of our method, we report visualization comparisons of retrieval results between our method, MS-SL \cite{MSSL}, and GMMFormer \cite{GMMFORMER} on ActivityNet Captions in Figure \ref{fig::retrieval_result}.
In the first case, MS-SL and GMMFormer superficially match videos with the action \textit{riding} and mistakenly associate \textit{riding} with \textit{bikes}, ignoring key semantics including \textit{pushes a rake} and \textit{scooter}.
In contrast, our method accurately retrieves the target video with correct semantics. The remaining example also shows a similar phenomenon, which demonstrates the robust and superior retrieval capability of our method.

\subsubsection{t-SNE Visualization}

We report t-SNE visualization \cite{tsne} to analyze the alignment of video and text features in the semantic space. For clearer observation, we randomly sample a subset of paired videos and text queries from Charades-STA. As shown in Figure \ref{fig:tsne} (a), in the base model, the distributions of video and text features exhibit limited overlap in the shared cross-modal space, indicating weak alignment between the two modalities. 
In contrast, our method achieves a more 
discriminative feature distribution.  As illustrated in Figure \ref{fig:tsne} (b), video and text features are tightly interleaved in regions, and the regions maintain sufficient separation to preserve semantics-specific distinctions. This suggests that our approach successfully aligns the two modalities while retaining their unique characteristics, leading to improved semantic alignment for PRVR.

\subsubsection{Text-Moment Similarity}
To verify the ability of our method to distinguish redundant moments and query-relevant moments, we present text-moment similarity examples on TVR. As shown in Figure \ref{fig:Text-Moment Similarity}, the base model produces similarity scores with limited fluctuations, indicating the weak distinction between distinct moment events. Conversely, our full model exhibits clear peak values, indicating enhanced discriminative ability in moment features. It suggests that our method effectively improves the ability to distinguish between query-relevant and query-irrelevant moments.

\section{Conclusion}
In this paper, we present a novel framework for Partially Relevant Video Retrieval, which systematically exploits the cross-model dual nature of video-text modalities: inter-sample correlation and intra-sample redundancy. 
Our approach introduces three key modules. 
The Inter Correlation Enhancement module strengthens cross-modal alignment by identifying and leveraging semantically similar yet unpaired text-video samples as pseudo-positive pairs.
The Intra Redundancy Mining module improves discriminative learning by distinguishing redundant moments and query-relevant moments, forcing the model to focus on query-relevant visual semantics.
The Temporal Coherence Prediction module enhances discrimination of fine-grained moment-level semantics through a self-supervised video sequence prediction task.
Extensive experiments on three datasets demonstrate that our framework achieves state-of-the-art performance.
{
    \small
    \bibliographystyle{ieeenat_fullname}
    \bibliography{main}
}

\end{document}